\documentclass[conference]{IEEEtran}
\usepackage{times}

\usepackage[numbers]{natbib}
\usepackage{multicol}
\usepackage[bookmarks=true]{hyperref}
\usepackage{booktabs}
\usepackage{amsmath}
\usepackage{amssymb}
\usepackage{graphicx}

\usepackage{tabularx}    % 自适应列宽
\usepackage{booktabs}    % 三线表（专业学术风格）
\usepackage{ragged2e}    % 文字左对齐自动换行
\usepackage{caption}      % 控制表格标题
\usepackage{booktabs}
\usepackage{multirow}
\usepackage{algorithm,algpseudocode}
\DeclareMathOperator{\Log}{Log}

\begin{document}

\title{Make Tracking Easy: Neural Motion Retargeting for 
Humanoid Whole-body Control}

% % You will get a Paper-ID when submitting a pdf file to the conference system
% \author{Qingrui Zhao$^{1}$, Kaiyue Yang$^{1}$, Xiyu Wang$^{1}$, Yi Lu$^{1}$, Xinfang Zhang$^{2}$, Wei Yin$^{3}$,\\ Qiu Shen$^{1}$, Xiao-Xiao Long$^{1}$, Xun Cao$^{1}$ 
% \thanks{$^{1}$Nanjing University, Nanjing, China.}
% \thanks{$^{2}$Huawei, China.}
% \thanks{$^{3}$Horizon Robotics, China.}}

% avoiding spaces at the end of the author lines is not a problem with
% conference papers because we don't use \thanks or \IEEEmembership

\author{
\authorblockN{
Qingrui Zhao\textsuperscript{1},
Kaiyue Yang\textsuperscript{1},
Xiyu Wang\textsuperscript{1, 2},
Shiqi Zhao\textsuperscript{1},
Yi Lu\textsuperscript{1},
Xinfang Zhang\textsuperscript{2}, \\
% Wei Yin\textsuperscript{3}, \\ 
Qiu Shen\textsuperscript{1},
Xiao-Xiao Long\textsuperscript{1*},
Xun Cao\textsuperscript{1*}
}
\authorblockA{\textsuperscript{*}Corresponding authors.}
\authorblockA{\textsuperscript{1}Nanjing University, \textsuperscript{2}Huawei Technologies}
\authorblockA{Project page: \url{https://nju3dv-humanoidgroup.github.io/nmr.github.io/}}
}

\maketitle

\begin{abstract}
Humanoid robots require diverse motor skills to integrate into complex environments, but bridging the kinematic and dynamic embodiment gap from human data remains a major bottleneck. We demonstrate through Hessian analysis that traditional optimization-based retargeting is inherently non-convex and prone to local optima, leading to physical artifacts like joint jumps and self-penetration. To address this, we reformulate the targeting problem as learning data distribution rather than optimizing optimal solutions, where we propose NMR, a Neural Motion Retargeting framework that transforms static geometric mapping into a dynamics-aware learned process.
We first propose Clustered-Expert Physics Refinement (CEPR), a hierarchical data pipeline that leverages VAE-based motion clustering to group heterogeneous movements into latent motifs. This strategy significantly reduces the computational overhead of massively parallel reinforcement learning experts, which project and repair noisy human demonstrations onto the robot’s feasible motion manifold. The resulting high-fidelity data supervises a non-autoregressive CNN-Transformer architecture that reasons over global temporal context to suppress reconstruction noise and bypass geometric traps. Experiments on the Unitree G1 humanoid across diverse dynamic tasks (e.g., martial arts, dancing) show that NMR eliminates joint jumps and significantly reduces self-collisions compared to state-of-the-art baselines. Furthermore, NMR-generated references accelerate the convergence of downstream whole-body control policies, establishing a scalable path for bridging the human-robot embodiment gap.

\end{abstract}

\IEEEpeerreviewmaketitle

\section{Introduction}

Humanoid robots are at a critical stage in their transition from laboratory settings to complex human environments, and the acquisition of diverse motor skills is fundamental to this progression. At present, the dominant research paradigm relies on large-scale human motion data, such as video recordings or motion-capture databases, as prior guidance for training robot motor control policies through imitation learning or reinforcement learning (RL)\cite{he2025hover, fu2024humanplus, cheng2024expressive}. Within this pipeline, motion retargeting serves as a critical bridge between human demonstrations and robotic execution. Conventional retargeting methods, including inverse kinematics (IK)-based approaches and differential optimization schemes such as GMR\cite{joao2025gmr}, primarily seek optimal joint configurations at the geometric level.

However, this conventional decoupled architecture of “retargeting first, tracking later” suffers from two major bottlenecks: \textbf{i) mathematical non-convexity}\cite{dai2019global, haviland2023manipulator, craig2009introduction}. Motion retargeting is inherently a highly non-convex optimization problem and is therefore prone to becoming trapped in local optima. As a result, such methods are highly sensitive to initialization and require tedious parameter tuning. When poorly initialized, they often produce physically infeasible artifacts, such as abrupt joint jerks, self-interpenetration, and foot sliding, thereby forcing downstream controllers to learn compensatory behaviors or lower stability. \textbf{ii) noise propagation.} Human motion data at the source side, such as SMPL-based estimations, often contain noise in the form of ground penetration or temporal jitter. Geometric optimization methods lack awareness of physical plausibility and therefore merely propagate these errors mechanically, leading to a classic “garbage in, garbage out” dilemma.

To address these limitations, we propose a Neural Motion Retargeting (NMR) framework. The central idea is to reformulate retargeting from static optimization over frame-wise states into dynamic mapping between motion distributions. In a data-driven manner, the model can directly learn a mapping from the human motion space to the robot’s feasible motion manifold. However, realizing this vision entails a chicken-and-egg dilemma: training a highly robust neural retargeter requires large-scale, high-quality robot motion data, yet such data are extremely difficult to obtain without an efficient retargeting tool. 

To obtain physical plausible motion data, we design a carefully structured hierarchical data pipeline, termed Clustering-Expert Physical Refinement (CEPR). We first use a variational autoencoder (VAE) to extract motion features and cluster heterogeneous human motion data accordingly. Subsequently, we train parallel reinforcement learning expert policies to drive the robot to track these clustered motion sets in a physics simulator, thereby automatically correcting physical defects in the original data and generating “ground-truth” motions that satisfy dynamic constraints. In summary, the contribution of this work includes: 

\begin{itemize}
    \item Proposed a Neural Motion Retargeting (NMR) framework for human to humanoid motion retargeting. By reformulating the retargeting problem as a distribution mapping from the human motion space to the robot’s motion manifold, our method alleviated issues in optimization-based methods such as local minima, joint discontinuities, and self-collisions.
    
    \item Introduced a hierarchical data construction pipeline, termed Clustered-Expert Physics Refinement (CEPR). Through motion clustering, parallel reinforcement learning expert policies, and physics-based simulation refinement, large-scale, high-fidelity, and physically consistent human–robot paired data are automatically generated, providing reliable supervision for neural retargeting model training.
    
    \item Developed a transformer-based motion retargeting network is and corresponding two-stage training strategy to enable broad motion coverage and bake physical feasibility.
    
    \item The effectiveness of the proposed method is validated through diverse motion experiments on the Unitree G1 robot. The results show that joint discontinuities, self-collisions, and joint-limit violations are significantly reduced by NMR, while the training efficiency and tracking performance of downstream whole-body control policies are improved.
\end{itemize}

% To further improve generalization, we adopt a two-stage training strategy. We first conduct large-scale pretraining on a vast set of preliminarily filtered geometrically retargeted data generated by GMR to establish a foundation model, and then fine-tune it on the physically consistent dataset produced by CEPR. 

% Experiments show that NMR not only eliminates the joint discontinuities commonly observed in conventional optimization-based methods, but also demonstrates the ability to correct upstream errors. Even when the input SMPL motion contains severe estimation errors or physically implausible artifacts, NMR can still transform it into a smooth reference trajectory that complies with robot dynamics.

% This advance not only simplifies the pipeline for robotic skill acquisition, but also provides a scalable path toward narrowing the embodied gap between humans and robots.

\section{Related Work}
Motion retargeting aims to transfer human motion data to humanoid robots while accounting for their distinct kinematic structures and physical constraints. We review related work from three perspectives: optimization-based retargeting methods, data-driven retargeting methods, and physics-based motion imitation.

\subsection{Optimization-based Retargeting Methods}

Motion retargeting originated from character animation research in computer graphics. Classical methods\cite{popovic1999physically, tak2005physically, lyard2008motion} employed optimization-based spacetime constraint solvers, formulating motion retargeting as optimization problems with kinematic constraints. While these approaches perform well for single-frame poses, they struggle to guarantee temporal consistency and physical feasibility.

In robotics, simple approaches\cite{cheng2024expressive, fu2025humanplus} directly copy joint rotations from human motion to the robot joint space. However, topological and morphological differences between humans and humanoid robots cause such direct mapping to produce artifacts including floating, foot penetration, and end-effector drift. To address joint space misalignment, Whole-Body Geometric Retargeting (WBGR) methods\cite{darvish2019whole, penco2018robust} IK to match Cartesian positions and orientations. However, these methods ignore human-robot scale differences and contact states, leading to floating, foot sliding, and ground penetration artifacts.

Recent advances incorporate parametric human body models. The PHC\cite{luo2023perpetual} leverages the SMPL model\cite{loper2023smpl} to fit robot skeleton shape parameters and solves IK through gradient descent, an approach widely adopted by H2O\cite{he2024learning}, HOVER\cite{he2025hover}, and OmniH2\cite{he2024omnih2o}. However, this method is computationally expensive and neglects contact constraints. GMR\cite{joao2025gmr} addresses these issues through non-uniform local scaling and two-stage IK optimization, significantly reducing foot sliding and self-penetration artifacts. PHUMA\cite{lee2025phuma} further introduces multiple physical constraints and jointly optimizes entire motion sequences.

Despite these advances, optimization-based methods remain fundamentally constrained by the inherent non-convexity of IK problems\cite{dai2019global, haviland2023manipulator, craig2009introduction, nocedal2006numerical}, leading to initialization sensitivity and frequent convergence to suboptimal solutions. This mathematical limitation motivates a shift toward data-driven paradigms.

\subsection{Data-driven Retargeting Methods}

To circumvent the local optima problem inherent in optimization methods, data-driven approaches in character animation achieve motion transfer by learning shared latent spaces across different skeletal structures, enabling cross-skeleton retargeting without paired data or 3D reconstruction \cite{villegas2018neural, aberman2020skeleton, lee2023same, yang2020transmomo}. These methods leverage cycle-consistency constraints, adversarial training, or contrastive learning to discover correspondences between different embodiments without explicit supervision\cite{hu2023pose, zhang2024semantics}.

In humanoid robotics, similar latent space alignment approaches have been explored for bridging the embodiment gap. Early efforts focused on learning shared representations for translating motions between humans and robots using manually collected paired datasets\cite{choi2020nonparametric}. To overcome the data bottleneck, recent works introduced self-supervised techniques for automating correspondence discovery: ImitationNet\cite{yan2023imitationnet} and its variants\cite{yagi2024unsupervised, stanley2021robust} employ GAN-based or cycle-consistency approaches for teleoperation without paired data. More recent work explores contrastive learning to enhance the expressiveness and smoothness of motion retargeting\cite{wang2024cross, yan2026learning}. However, these methods primarily focus on upper-body manipulation or simple arm motions due to the difficulty of acquiring paired whole-body data and satisfying dynamic constraints for locomotion.

Existing data-driven methods face two limitations: they inherit local optima artifacts from optimization-based supervision, and lack physical reasoning to filter out source motion noise such as ground penetration and temporal jitter.

\section{Method}

We propose a neural motion retargeting framework that learns a direct mapping from human SMPL motion sequences to feasible humanoid robot motion, bypassing the local-optima failures commonly observed in conventional optimisation-based approaches. 
\begin{figure*}[htb]
	\centering
	\includegraphics[width=\linewidth]{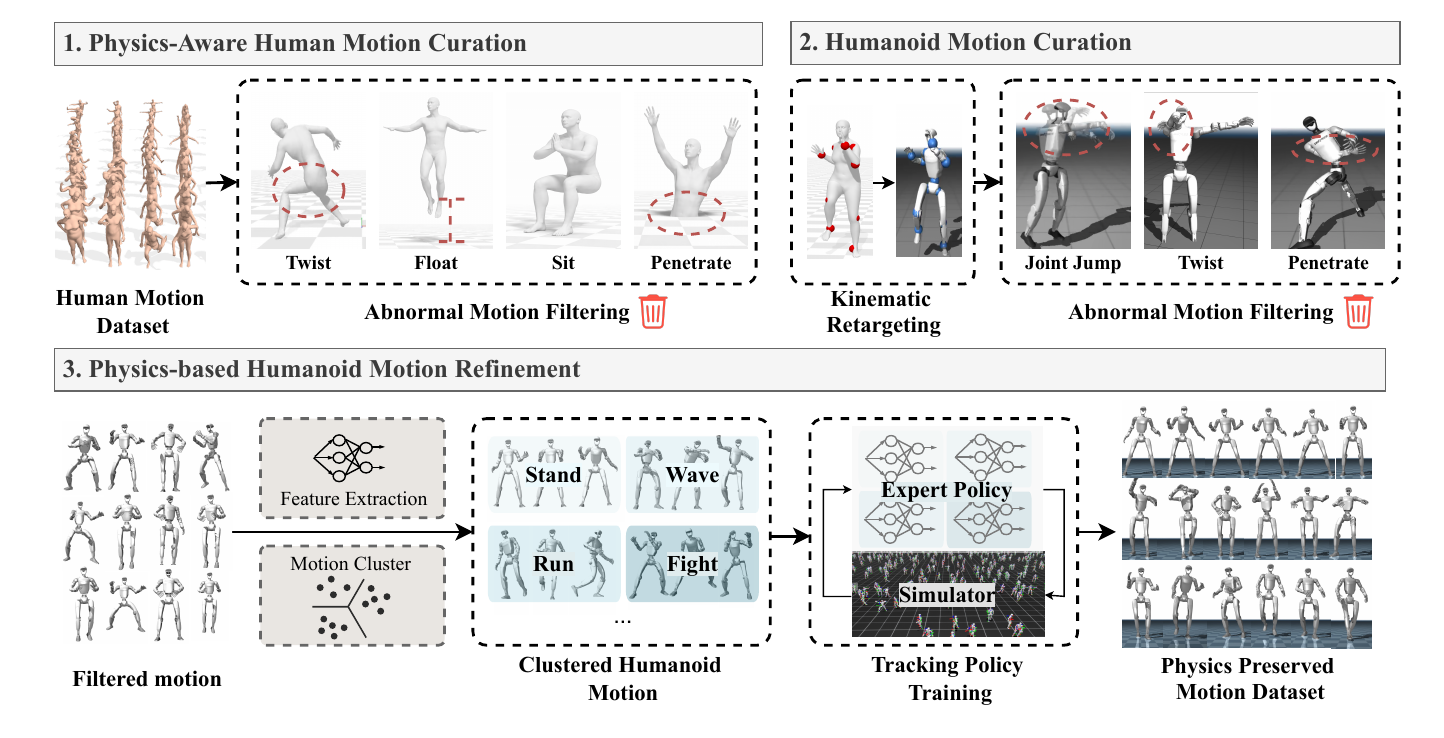}
	\caption{Data Construction Pipeline. We obtain high-quality human–humanoid motion pairs through three processing stages.}
    \label{fig:saf pipeline}
\end{figure*}

In Section \ref{sec: problem_formulation}, we introduce typical optimization-based retargeting methods to provide a non-convexity analysis and explained why these methods frequently stall in local optima. This theoretical limitation motivates a shift toward our data-driven paradigm that can circumvent these geometric traps. In Section \ref{sec:dataset}, we propose a three-stage "Clustered-Expert Physics Refinement" pipeline, which yields approximately 30,000 physics preserved SMPL-robot motion dataset. In Section~\ref{sec:network}, we propose a motion retargeting network that learns to directly map human SMPL sequences to physically feasible humanoid motion, and can  suppress artifacts caused by upstream pose. In Section \ref{sec:training scheme}, we introduce our ``Detail to physical'' training scheme that can guarantee both retargeted motion quality and physical fidelity.  

\subsection{Preliminary: Optimization-based Human to Humanoid Retargeting} ~\label{sec: problem_formulation}

\noindent\textbf{Optimization-based Retargeting Methods. }

Motion retargeting aims to convert a human SMPL motion sequence $\{\textbf{s}_t\}^{T}_{t=1}$ into a robot-executable sequence $\{\textbf{q}_t\}^{T}_{t=1}$, where $\textbf{q}$ is robot generalized coordinates (root translation, root rotation, and joint values), while preserving motion semantics and satisfying the robot's physical constraints. Conventional approaches such as GMR\cite{joao2025gmr} formulate this as a per-frame optimization problem:
\begin{equation}
\begin{split}
    f(\textbf{q}) = \sum_{(i,j)\in\mathcal{M}} w^R_{i,j} \|R^h_i \ominus R_j(\textbf{q})\|^2 \\
           + \sum_{(i,j)\in\mathcal{M}_{ee}} w^p_{i,j}
             \|p^{\mathrm{target}}_i - p_j(\textbf{q})\|^2
\end{split}
\end{equation}

\noindent where $R_i^h \in S O(3)$ is the orientation of the human body $i$, $\mathbf{p}_j(\mathbf{q})$ and $R_j(\mathbf{q}) \in S O(3)$ are the Cartesian position and orientation of the robot body $j$, $\textbf{q}$ is robot generalized coordinates (root translation, root rotation, and joint values), $R_i \ominus R_j$ is the exponential map representation of the orientation difference between $R_i$ and $R_j$. Although this formulation is formally concise, the underlying optimization is highly non-convex, owing to two mutually coupled geometric challenges.

\noindent\textbf{Non-convexity Analysis. } 

For notational clarity in the following analysis, we denote the joint configuration as $\boldsymbol{\theta} \in \mathbb{R}^n$, which corresponds to the joint-angle components of $\textbf{q}$.
To expose the geometric source of non-convexity, we consider a single end-effector pose $T(\boldsymbol{\theta})\in SE(3)$ and a target pose $T^\star\in SE(3)$, and define the relative pose error $E(\boldsymbol{\theta}) = (T^\star)^{-1}T(\boldsymbol{\theta}) \in SE(3)$. We lift this error to the Lie algebra via
\begin{equation}
    \boldsymbol{\xi}(\boldsymbol{\theta})
    = \Log(E(\boldsymbol{\theta}))
    = \begin{pmatrix}
        \boldsymbol{\omega}(\boldsymbol{\theta}) \\
        \mathbf{v}(\boldsymbol{\theta})
      \end{pmatrix}
    \in \mathbb{R}^6,
    \label{eq:xi}
\end{equation}
where $\boldsymbol{\omega}\in\mathbb{R}^3$ is the rotational log-coordinate and $\mathbf{v}\in\mathbb{R}^3$ is the translational log-coordinate. We then study the weighted surrogate cost
\begin{equation}
    \tilde f(\boldsymbol{\theta})
    = \frac{1}{2}\,\boldsymbol{\xi}(\boldsymbol{\theta})^\top
    W\,
    \boldsymbol{\xi}(\boldsymbol{\theta}),
    \qquad
    W=\mathrm{diag}(w_R I_3,\; w_p I_3),
    \label{eq:surrogate_cost}
\end{equation}
with weights $w_R,w_p>0$. By direct differentiation, the Hessian of this surrogate objective decomposes as
\begin{equation}
    \nabla^2 \tilde f(\boldsymbol{\theta})
    =
    \underbrace{J_{\xi}^\top W J_{\xi}}_{\text{Gauss--Newton term, PSD}}
    +
    \underbrace{\sum_{a=1}^{6}(W\boldsymbol{\xi})_a \,\nabla^2 \xi_a}_{\text{curvature correction}},
    \label{eq:hessian_decomp}
\end{equation}
where $J_{\xi} = \frac{\partial \boldsymbol{\xi}}{\partial \boldsymbol{\theta}} \in \mathbb{R}^{6\times n}$ is the Jacobian of the log-coordinate error. The first term is always positive semi-definite; the negative curvature arise from the second term, which captures the second-order geometry of the error map $\boldsymbol{\theta}\mapsto \Log((T^\star)^{-1}T(\boldsymbol{\theta}))$.

Detailed analysis (see Appendix~\ref{app:nonconvexity}) reveals two geometrically distinct sources of non-convexity: (i) the curvature of forward kinematics due to recursive trigonometric composition, and (ii) the nonlinearity of the logarithmic map on $SO(3)$/$SE(3)$. Together they imply:

Proposition 1 (existence of negative curvature).
\emph{
Let $n\ge 2$ and $w_R,w_p>0$. For the surrogate objective $\tilde f$ in~\eqref{eq:surrogate_cost}, there exist feasible target poses $T^\star$ and robot configurations $\boldsymbol{\theta}$ such that the Hessian $\nabla^2 \tilde f(\boldsymbol{\theta})$ has a strictly negative directional curvature; namely, there exists a direction $\mathbf{u}\neq \mathbf{0}$ satisfying
\begin{equation}
    \mathbf{u}^\top \nabla^2 \tilde f(\boldsymbol{\theta}) \mathbf{u}<0.
\end{equation}
Hence, $\tilde f$ is generally non-convex.
}

Proposition~1 establishes that the retargeting landscape admits configurations with strictly negative curvature generated by either forward-kinematics curvature or logarithmic-map curvature. This explains why gradient-based retargeting can be sensitive to initialization and may stall in poor local minima even for seemingly simple target motions.

\noindent\textbf{Our Core Idea.}

The non-convexity of retargeting creates practical limitations. Differential Inverse Kinematics (Differential IK) linearizes the objective at the current iterate, reducing the problem to a convex quadratic program. However, this approximation tends to be valid only within a limited basin of attraction around the true solution, offers no assurance of global convergence, and can be sensitive to initialization and the choice of weighting parameters.

We reframe retargeting as a supervised learning task. But this shift is difficult for two reasons: (i) data: naive use of kinematic retargeting outputs as supervision would inherit the same local-optima failures, making high-quality training pairs difficult to obtain; and (ii) architecture: the network must generalize across diverse motion styles while respecting physical feasibility. To address these challanges, we introduce physics-simulation-based data generation pipeline (\ref{sec:dataset}) and motion retargeting network (\ref{sec:network}). 

\subsection{Clustered-Expert Physics Refinement.} ~\label{sec:dataset}

The systematic transformation from raw human motion to robot-feasible trajectories is summarized in Figure \ref{fig:saf pipeline}. This pipeline acts as a hierarchical filter: Steo 1 ensures semantic relevance, Step 2 enforces kinematic validity, and Step 3 leverages the physics engine to resolve dynamic inconsistencies, eventually yielding a high-fidelity dataset for neural network supervision.

\noindent\textbf{Step 1 Physics-Aware Human Motion Curation. } 

The raw SMPL dataset contains a large number of motions that are semantically incompatible with robotic applications that would introduce substantial spurious noise. Following PHUMA\cite{lee2025phuma}, we apply the same filtering method to remove physically inconsistent motion including (i) excessive jerk, (ii) a CoM position far outside its support base, or (iii) insufficient foot-ground contact (float and penetration).

After this stage, all retained sequences correspond semantically to motions that the robot can in principle execute, thereby providing a foundation for the subsequent fine-grained filtering steps.

\noindent\textbf{Step 2 Kinematic Retargeting and Quality Filtering.}

% 修改，简化，主要说这是我怎么做的
Then, we employ optimization-based kinematic retargeting method (GMR\cite{joao2025gmr}) on curated SMPL sequences to obtain the initial corresponding robot motion dataset. While GMR effectively mitigates common artifacts via human-robot rest-pose alignment and multi-stage inverse-kinematics (IK) optimization, it remains fundamentally constrained by the non-convex nature of the optimization problem (as analyzed in Section \ref{sec: problem_formulation}). Consequently, the output quality cannot be universally guaranteed, particularly for complex or highly dynamic motions.

To address potential optimization failures or IK singularities, we apply a hard-threshold filtering pipeline. A motion segment is preserved only if it satisfies the following joint-space and geometric constraints:

\begin{itemize}
    \item Joint Continuity and Stability: We compute the inter-frame joint velocity $\dot{\mathbf{q}}_t = (\mathbf{q}_t - \mathbf{q}_{t-1})/\Delta t$. Any segment containing a peak velocity $|\dot{\mathbf{q}}_t| > \dot{\mathbf{q}}_{max}$ is discarded, where $\dot{\mathbf{q}}_{max}$ represents the hardware-specific velocity saturation limit. This prunes abrupt, non-smooth jumps caused by IK singularities.
    \item Geometric Self-Intersection Detection: We leverage the MuJoCo collision detection engine to identify interpenetrations of the robot's geometric links. By loading the robot's URDF into a simulation context, we check for contacts between all geoms. A sequence is rejected if the fraction of self-intersecting frames exceeds a tolerance $\text{cross\_ratio} = 0.05$.
    \item Floating Foot Rectification: To ensure the motion is physically grounded, we calculate the average foot clearance relative to the estimated ground plane. If the mean elevation of the lowest foot point across the sequence exceeds $\text{float\_threshold} = 0.10$ m, the motion is classified as "floating" (e.g., sitting or lying poses in the air) and pruned from the training set.
\end{itemize}

\noindent \textbf{Step 3 Physics-based Humanoid Motion Refinement}\label{sec: rl physics refinement}

This stage constitutes the core of the entire data pipeline, with the objective of transforming kinematic references into physically consistent robot motion trajectories, using the physics simulator and RL policy as the presudo ground-truth source.

% 聚类SMPL动作
\vspace{1em}

\noindent \textit{ Motion Clustering.}  

Training a single RL tracking policy over the full motion dataset suffers from distributional conflict \cite{luo2023perpetual}, leading to unstable performance and degraded tracking accuracy. Conversely, training a separate policy on individual motion sequences is prohibitively expensive in both computation and training time. To address this, we partition the motion library into behaviorally related clusters, allowing each expert policy to specialize over a homogeneous motion distribution.

Specifically, we leverage TMR \cite{petrovich2023tmr} to train a motion-text retrieval model via contrastive and reconstruction losses, which establishes a well-structured cross-modal latent space. The resulting motion encoder is then used to extract latent representations for all motion sequences. Our goal is to cluster motions by semantic type, for instance, grouping jumps into one cluster and in-place motions into another. The semantically-aligned motion-text features ensure that motions sharing similar semantics are embedded in proximity, even if their kinematic patterns differ. We then apply K-Means algorithm with cosine similarity as the distance metric to partition all motion sequences in the latent space.

% Training a single RL tracking policy across the full motion library suffers from distributional conflict\cite{luo2023perpetual}, resulting in unstable performance and lower tracking accuracy. Therefore, we partition the motion library into behaviorally coherent clusters, so that each expert policy operates over a sufficiently homogeneous distribution and can specialize in the control demands of its assigned motion type.

\vspace{1em}
\noindent \textit{Expert Policy Training.} 

For each cluster, we train an RL tracking policy in a massively parallelized physics simulation environment to make the policy to replicate the target motion in simulator as close as possible. We leveraged symmetric actor-critic framework and PPO algorithm for policy training to maximize the data effency. 
\begin{table}[t]
\centering
\caption{Policy Observation Space}
\label{tab:observations}
\begin{tabular}{llc}
\toprule
\textbf{Observation term} & \textbf{Description} & \textbf{Dimension} \\
\midrule
\multicolumn{3}{l}{\textit{Reference motion state} $s_t^g$} \\
\quad $\boldsymbol{q}^g$ & Reference joint positions & 29 \\
\quad $\dot{\boldsymbol{q}}^g$ & Reference joint velocities & 29 \\
\quad $\boldsymbol{p}_{b}^g$ & Reference body positions (world) & $14 \times 3 = 42$ \\
\quad $\boldsymbol{v}_{b}^g$ & Reference body linear velocities & $14 \times 3 = 42$ \\
\quad $\boldsymbol{o}_{b}^g$ & Reference body orientations (quat) & $14 \times 4 = 56$ \\
\quad $\boldsymbol{\omega}_{b}^g$ & Reference body angular velocities & $14 \times 3 = 42$ \\
\midrule
\multicolumn{3}{l}{\textit{Robot proprioception} $s_t^p$} \\
\quad $\boldsymbol{p}_{b}^p$ & Robot body positions (world) & $14 \times 3 = 42$ \\
\quad $\boldsymbol{v}_{b}^p$ & Robot body linear velocities & $14 \times 3 = 42$ \\
\quad $\boldsymbol{o}_{b}^p$ & Robot body orientations (quat) & $14 \times 4 = 56$ \\
\quad $\boldsymbol{\omega}_{b}^p$ & Robot body angular velocities & $14 \times 3 = 42$ \\
\quad $\boldsymbol{q}^p$ & Robot joint positions (relative) & 29 \\
\quad $\dot{\boldsymbol{q}}^p$ & Robot joint velocities (relative) & 29 \\
\midrule
\multicolumn{3}{l}{\textit{Action history}} \\
\quad $\boldsymbol{a}_{t-1}$ & Previous action & 29 \\
\midrule
& \textbf{Total} & \textbf{509} \\
\bottomrule
\end{tabular}
\end{table}

As shown in Table~\ref{tab:observations}, we provide the policy with comprehensive state information including both the reference motion state $s_t^g$ and the robot's proprioceptive state $s_t^p$. This rich observation space enables the policy to accurately perceive the tracking error between the current robot configuration and the target motion, facilitating precise whole-body motion tracking.

As shown in Table~\ref{tab:rewards}, we mainly add tracking rewards and only minimal regulation rewards to guarantee accurate motion tracking. Moreover, we employ an adaptive standard deviation schedule for the tracking rewards. When training on large-scale motion datasets, the effective sample count per motion decreases compared to single-motion training. To compensate for this reduced sample efficiency and achieve lower tracking errors, we progressively tighten the reward function by decreasing $\sigma$ from $\sigma_{\text{start}}$ to $\sigma_{\text{end}}$ over the course of training:
\begin{equation}
    \sigma(i) = \sigma_{\text{start}} + (\sigma_{\text{end}} - \sigma_{\text{start}}) \cdot \frac{i - i_0}{i_{\max} - i_0}
\end{equation}
where $i$ denotes the current training iteration. This curriculum learning strategy allows the policy to first learn coarse motion patterns with relaxed reward tolerances, then gradually refine the tracking precision as training progresses.
\begin{table}[t]
\centering
\caption{Reward Terms for Expert policy training}
\label{tab:rewards}
\begin{tabular}{lc}
\toprule
\textbf{Reward term} & \textbf{Weight} \\
\midrule
\multicolumn{2}{l}{\textit{Tracking rewards} $\mathcal{R}(s_t^p, s_t^g)$} \\
\quad Anchor position & 1.0 \\
\quad Anchor orientation & 1.0 \\
\quad Anchor velocity & 1.0 \\
\quad Body link position (rel.) & 1.0 \\
\quad Body link orientation (rel.) & 1.0 \\
\quad Body link linear velocity & 1.0 \\
\quad Body link angular velocity & 1.0 \\
\midrule
\multicolumn{2}{l}{\textit{Penalty terms} $\mathcal{P}(s_t^p, a_t)$} \\
\quad Action rate & -0.1 \\
\quad Undesired contacts & -0.1 \\
\bottomrule
\end{tabular}
\end{table}

\vspace{1em}
\noindent \textit{ Generating Physically Faithful Data Pairs.} 

\begin{figure}[t]
	\centering
	\includegraphics[width=0.9\linewidth]{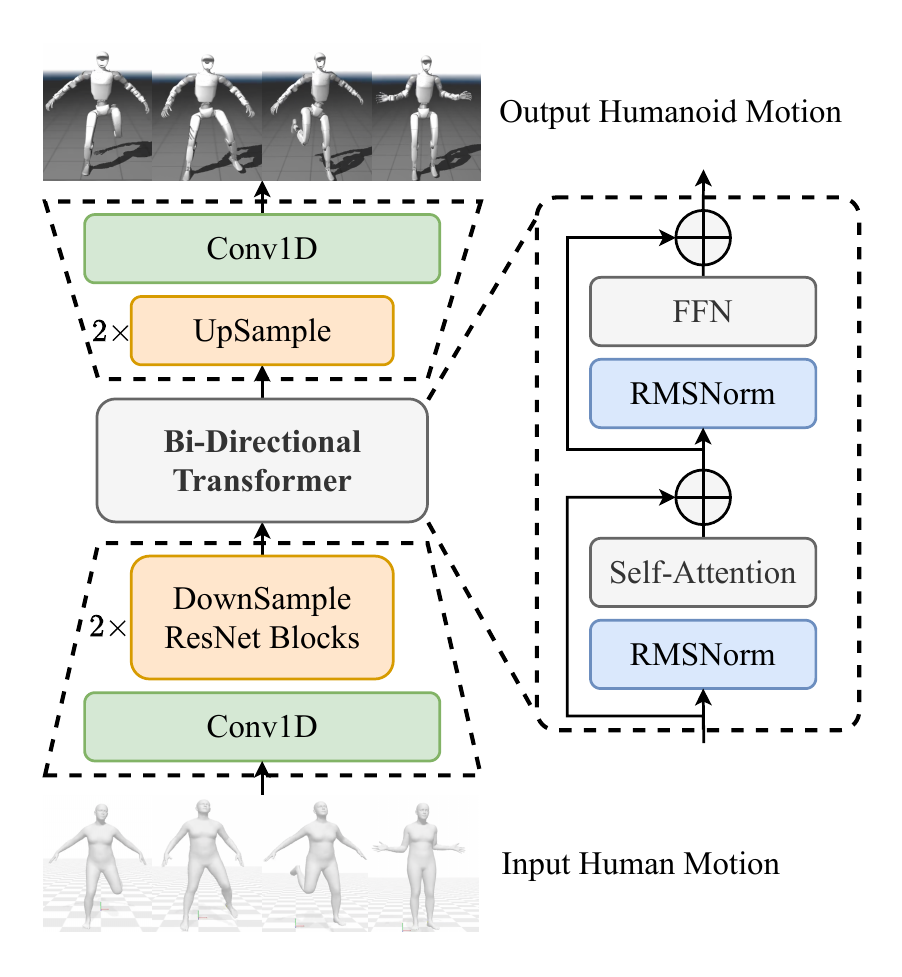}
	\caption{Overview of our neural motion retargeting network, which maps human motion to humanoid motion.}
    \label{fig:network}
\end{figure}  

\begin{figure*}[t]
	\centering
	\includegraphics[width=\linewidth]{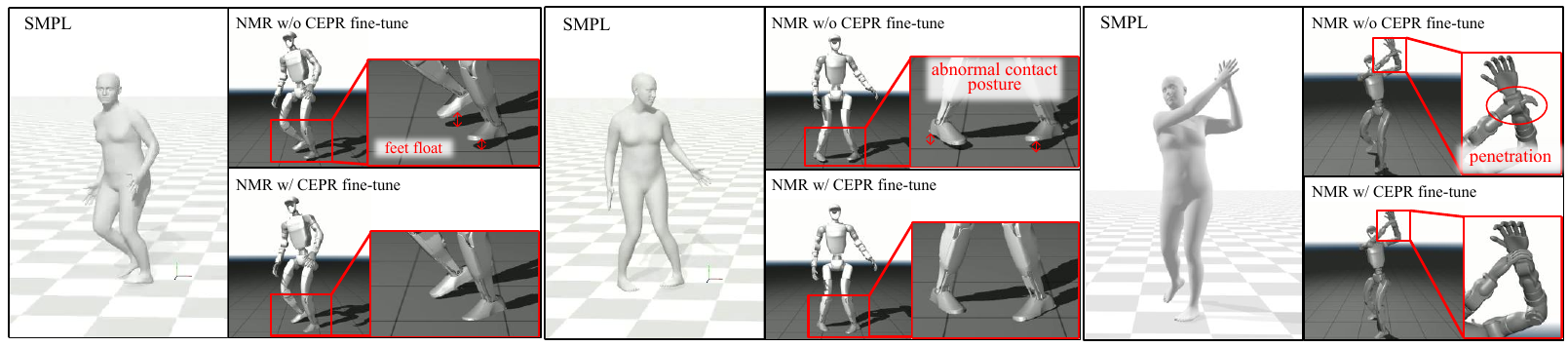}
	\caption{Visualization of NMR retargeting results with and without CEPR data fine-tuning}
\end{figure*}

Once the expert policies have converged, we roll out each policy in the simulator over its corresponding reference sequences, record the resulting robot state trajectories, and pair them with the corresponding input SMPL sequences, forming (SMPL sequence, physically consistent robot motion) data pairs. This process yields approximately 30K paired sequences in total, each implicitly validated by the physics simulation, constituting a compact yet high-quality subset of our overall dataset.

\subsection{Motion Retargeting Network} ~\label{sec:network}

\noindent\textbf{Motion Representation.}

Human and humanoid motion differ in joint structure, including the number of joints and the fact that humanoid robots are actuated by DoFs rather than full joint rotations. To accommodate these differences while maintaining alignment, we adopt distinct motion representations for humans and humanoids.

For human motion, we build upon the 272-dimensional representation used in MotionMillion \cite{fan2025gotozero} and reformulate it as:
\begin{equation}
    m^i = \{ r^{x}, r^{z}, r, j^{p}, j^{v} \},
\end{equation}
where $r^{x}, r^{z} \in \mathbb{R}$ denote the root linear velocities on the XZ-plane, $r \in \mathbb{R}^6$ represents the root orientation in 6D rotation representation, and $j^{p}, j^{v} \in \mathbb{R}^{3k}$ denote local joint positions and velocities, respectively. 
Humanoid motion is represented similarly, with the addition of the robot's joint DoFs:
\begin{equation}
    m_{bot}^i = \{ r_{\text{bot}}^{x}, r_{\text{bot}}^{z}, r_{\text{bot}}, j_{\text{bot}}^{p}, j_{\text{bot}}^{v}, q \},
\end{equation}
where $r_{\text{bot}}^{x}$ and $r_{\text{bot}}^{z}$ denote the humanoid root linear velocities on the XZ-plane, $r_{\text{bot}}$ denotes the root orientation, $j_{\text{bot}}^{p}, j_{\text{bot}}^{v} \in \mathbb{R}^{3d}$ denote local joint positions and velocities, and $q \in \mathbb{R}^{n}$ denotes the joint DoFs.

\noindent\textbf{Motion Retargeting Network.}

As illustrated in Fig.~\ref{fig:network}, our neural motion retargeting network directly maps human motion to corresponding humanoid sequences. Given a human motion sequence, we first extract latent features via a 1D ResNet-based encoder. The features are then passed into a Transformer-based network following the architectural design of LLaMA~\cite{touvron2023llama}. Since human and humanoid motions are temporally aligned with a strict one-to-one correspondence, we replace causal attention with full self-attention, enabling parallel timestep-wise prediction conditioned on the entire input sequence. The output of the Transformer is subsequently decoded through upsampling and 1D-Conv layers to produce the final humanoid sequences. The network is optimized by minimizing an L1 loss:
\begin{equation}
\mathcal{L} = \sum_{t=1}^{T} \|m_{\text{bot}}^t - \hat{m}_{\text{bot}}^t\|_1,
\end{equation}
where T denotes the length of the motion sequence.

% Physics Injected Training Scheme
  \subsection{Two-Stage Training Scheme} ~\label{sec:training scheme}

  % [Opening — 动机]
  Corresponding to the two-tier data hierarchy described in Section~\ref{sec:dataset},
  we train the retargeting network in two sequential stages: large-scale kinematic
  pre-training followed by physics-guided fine-tuning.
  This strategy is necessary because neither dataset alone is sufficient: the kinematic
  dataset provides breadth but lacks physical guarantees, while the physics dataset
  provides feasibility signals but is too small to support generalization.

 % [Stage 1]
\noindent \textbf{Step 1: Kinematic Alignment with large-scale data.}

  We first pre-train the network on the large-scale kinematic retargeting dataset,  minimizing the regression loss between the predicted G1 joint angles and the kinematic reference targets. Despite the residual physical artifacts in kinematic data (e.g., foot skating, ground penetration), the dataset's scale and diversity endow the network with
  a foundational embodiment mapping across a broad range of motion categories, including locomotion, upper-limb manipulation, and martial-arts motions.

  % [Stage 2]
  \noindent \textbf{Step 2: Physical Grounding with CEPR data.}
  
  Building on the pre-trained checkpoint, we fine-tune the network using approximately  30{,}000 physically consistent motion pairs validated by physics simulation (Section~\ref{sec: rl physics refinement}). Although this dataset is smaller than the kinematic set by roughly an order of
  magnitude, each sample carries a strong physical feasibility signal that
  has been verified through RL policy rollouts in simulation. Fine-tuning shifts the output distribution toward the robot's dynamically feasible
  motion manifold, enabling the network to implicitly suppress physically infeasible components in upstream SMPL-X noise rather than propagating them to the output.

  % [Necessity argument]
  The necessity of both stages can be argued from two directions.
  Pre-training without physics fine-tuning (the \textit{NMR w/o RL} ablation) produces kinematically plausible but physically unconstrained outputs. Conversely, training on physics data alone without pre-training leads to overfitting to the limited set of motion patterns covered by the RL training corpus, failing to generalize to unseen motion types. Together, the two stages equip the network with both broad motion coverage and improved physical feasibility.

\section{Experiment}

\subsection{Experiment Setup}

% For the training of motion retargeting network, during pre-training phase,  we use the AdamW optimizer with a batch size of 128 per GPU and an initial learning rate of $2\times10^{-4}$, scheduled by cosine annealing. The network is trained for 500 epochs on 8 NVIDIA H20 GPUs. During fine-tuning phase, we used the same optimizer and batch size as pre-training phase, only reduced learning rate to $ 1\times 10^{-5}$ for warm start and trained for 50 epochs on 8 NVIDIA 4090 GPUs.

For training the motion retargeting network, we adopt a two-stage optimization scheme. During the kinematic-alignment setp, we use the AdamW optimizer with a batch size of 128  and an initial learning rate of $2 \times 10^{-4}$, scheduled via cosine annealing. The network is trained for 500 epochs.
During the physical grounding step, we use the same optimizer and batch size, while reducing the learning rate to $1 \times 10^{-5}$ for a warm start, and train for an additional 50 epochs.

\subsection{Datasets and Baselines}
To evaluate the proposed model's performance in human-to-robot retargeting, we curated a test suite from the AMASS\cite{AMASS:ICCV:2019} dataset containing 82 motion sequences (totaling 119K frames at 120Hz). All test data were strictly excluded from the training phase to ensure unbiased assessment. The dataset is categorized by Motion Complexity and Sequence Length:

\subsubsection{Motion Complexity} 
Sequences are classified into three levels based on kinematic mapping and hardware constraints:
\begin{itemize}
    \item Upper-limb-only (ULOM): Involves stationary lower bodies; evaluates workspace mapping and self-collision avoidance.
    \item Whole-body primitive (WBPM): Includes basic locomotion (e.g., walking, running); assesses coordinated multi-joint retargeting and Center of Mass (CoM) stability.
    \item Whole-body complex (WBCM): Covers high-dynamic motions (e.g., acrobatics, martial arts); tests robustness against mechanical joint limits and singularities.
\end{itemize}

\subsubsection{Sequence Length} 
To measure temporal stability and error accumulation, data is partitioned by frame count:
\begin{itemize}
    \item Short ($<$ 250 frames): Evaluates initialization speed and transient response to sudden kinematic changes.
    \item Medium (250--1000 frames): Assesses motion smoothness and the consistency of cyclic gait patterns.
    \item Long ($>$ 1000 frames): Tests the suppression of cumulative errors and drift over extended operations.
\end{itemize}

We adopt GMR\cite{joao2025gmr} as baseline methods. GMR employs a unique non-uniform scaling strategy and a two-stage optimization scheme to maximize the likelihood of obtaining a good initialization for the inverse kinematics optimization problem, thereby enabling accurate motion mapping. PHUMA improves the physical plausibility of the retargeted motions by introducing multiple physical constraints and jointly optimizing the entire motion sequence.

We use BeyondMimic\cite{liao2025beyondmimic} for RL-based tracking policy training. We keep reward, policy observation, and domain randomization settings identical to the Beyondmimic, only set $\mathrm{num\_envs}=16384 $ to improve value sampling. 

\subsection{Retargeting Quality and Policy Tracking}
% 定量检测retarget出来的motion中

\begin{figure}[t]
	\centering
	\includegraphics[width=\linewidth]{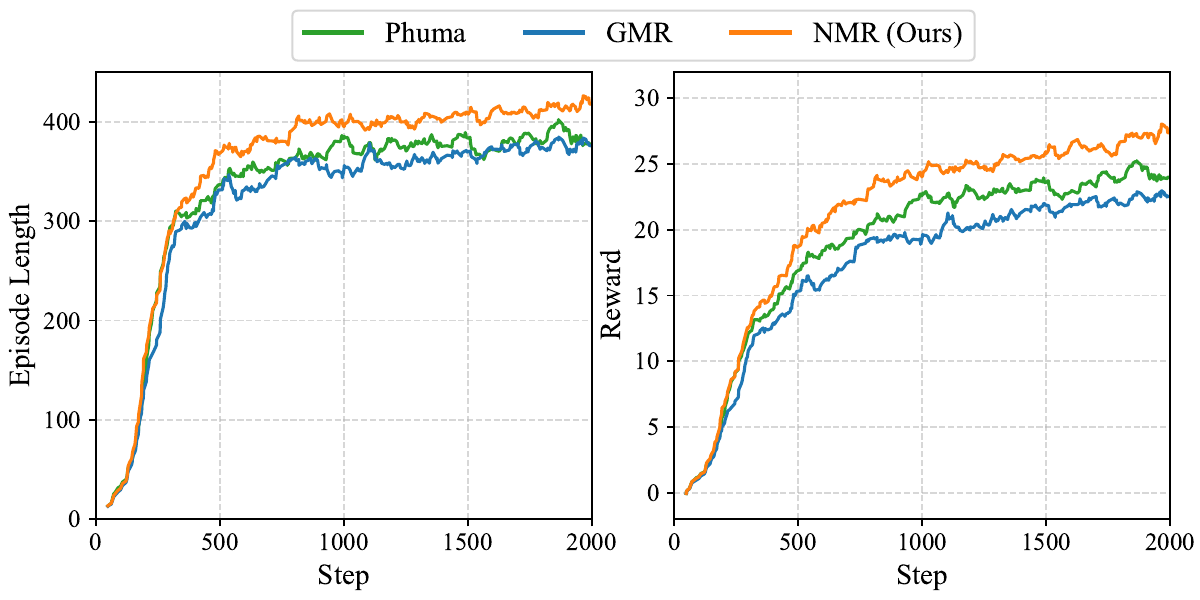}
	\caption{Comparison of training episode length and reward of motion retargeted with different methods.}
     \label{fig:train_log}
\end{figure}

\begin{table}[t]
\centering
\caption{Quantitative comparison of different methods in terms of joint jump, self collision and reaching joint limit}
\setlength{\tabcolsep}{4pt}
\small
\begin{tabular}{lccc}
\toprule
Method & Joint Jump $\downarrow$ & Self Collision $\downarrow$ & Joint Limit $\downarrow$ \\
\midrule
GMR           & 56 (0.11\%) & 947 (1.91\%) & 21443 (43.21\%) \\
PHUMA         & 12 (0.02\%) & 2456 (4.95\%) & 10580 (21.32\%) \\
NMR w/o RL & 2 (0.00\%)  & 779 (1.57\%) & 19324 (38.94\%) \\
\textbf{NMR }       &\textbf{ 0 (0.00\%)}  & \textbf{431 (0.87\%)} & \textbf{8339 (16.80\%)} \\
\bottomrule
\end{tabular}
\label{tab:quant_cmp}
\end{table}

%%%%%%%%%%%% 跨页大图 %%%%%%%%%%%% 但是信息量太低，用小图替代了
% \begin{figure*}[t]
%     \centering
%     \includegraphics[width=\textwidth]{fig/visual_cmp_1_motion.pdf}\par
%     \vspace{-1mm} % 调小两张图之间的竖向间距，可改成 -2mm, 0mm 等试

%     \includegraphics[width=\textwidth]{fig/visual_cmp_1_joint.pdf}
%     \caption{Visual comparison of retargeting result between PHUMA, GMR and NMR (Ours). PHUMA exhibit abnormal bend in waist roll direction and deviates the resulting motion from original SMPL model.}
%     \label{fig:visual cmp 1}
% \end{figure*}
%%%%%%%%%%%% 跨页大图 %%%%%%%%%%%%

\begin{figure}[t]
    \centering
    \includegraphics[width=\linewidth]{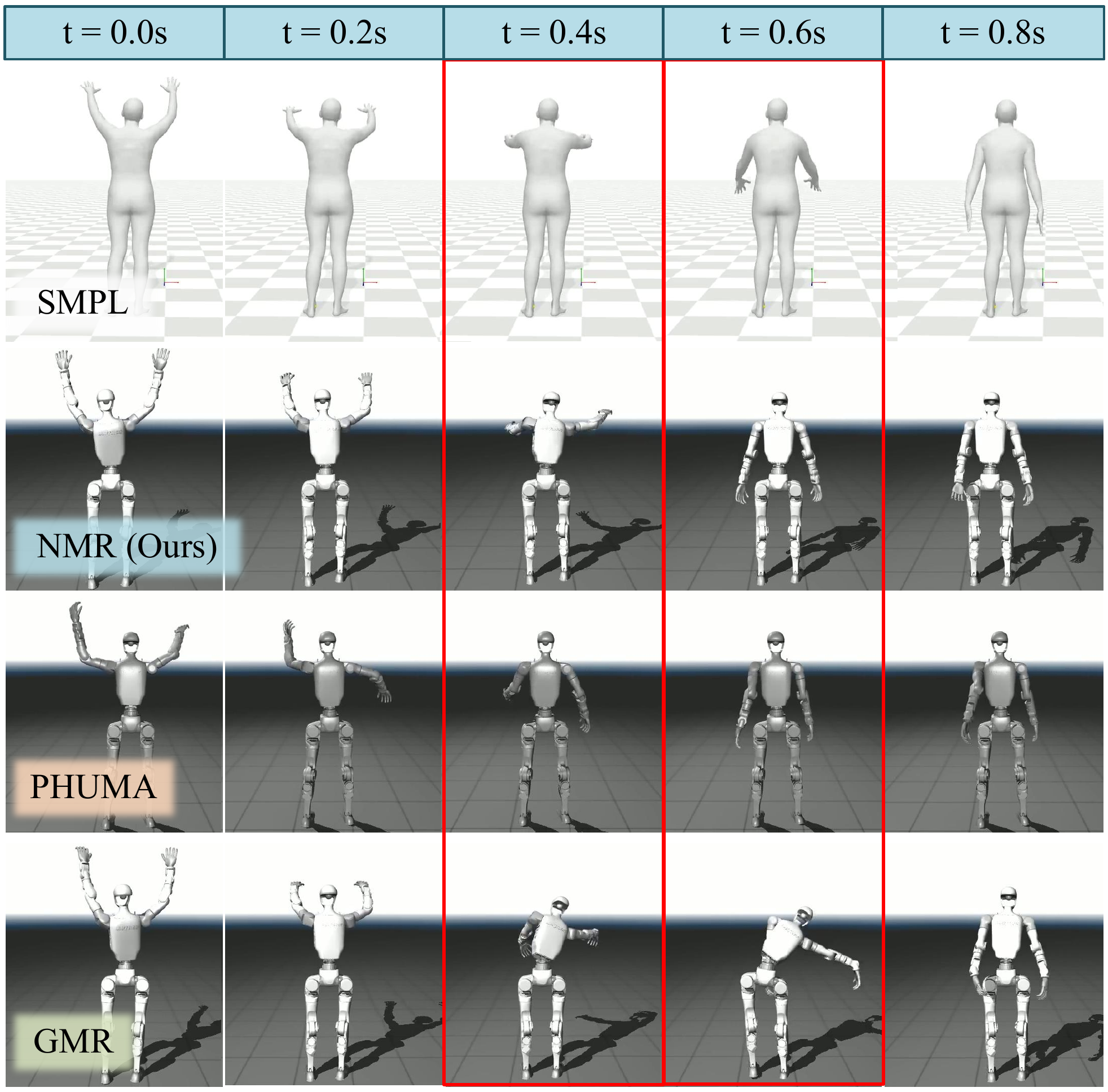}\par
    \vspace{0mm} % 调小两张图之间的竖向间距，可改成 -2mm, 0mm 等试
    \includegraphics[width=\linewidth]{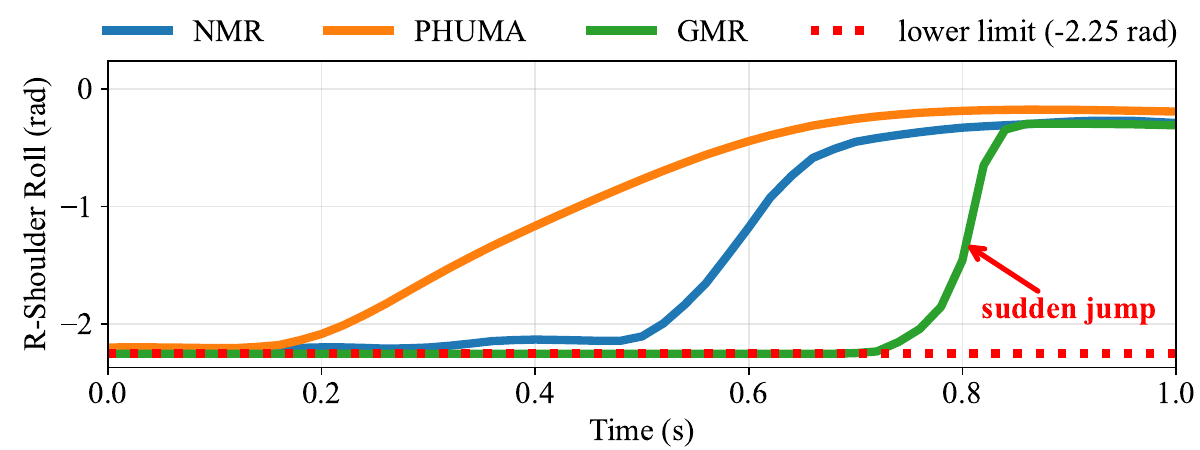}
    \caption{Visual comparative of different motion retargeting methods when processing a human "arm raise-lower" motion sequence. The top row displays the reference motion sequence from the SMPL human body model, while the subsequent three rows illustrate the retargeted robot motions generated by NMR (ours), PHUMA, and GMR, respectively. The red bounding boxes highlight the key frames where GMR exhibits significant motion anomalies. The line graph at the bottom illustrated ``right shoulder roll angle" of the retargeted motion. }
    \label{fig:visual cmp 1}
\end{figure}

  \begin{table*}[tp!]
  \centering
  \caption{Comparison of tracking accuracy. }
  \label{tab:tracking}
  \begin{tabular}{lccccccccc}
    \toprule
    \multirow{2}{*}{Method}
      & \multicolumn{3}{c}{Short}
      & \multicolumn{3}{c}{Medium}
      & \multicolumn{3}{c}{Long} \\
    \cmidrule(lr){2-4} \cmidrule(lr){5-7} \cmidrule(lr){8-10}
      & Success Rate$\uparrow$ & MPJPE $\downarrow$ & W-MPJPE $\downarrow$
      & Success Rate$\uparrow$ & MPJPE $\downarrow$ & W-MPJPE $\downarrow$
      & Success Rate$\uparrow$ & MPJPE $\downarrow$ & W-MPJPE $\downarrow$ \\
    \midrule
    PHUMA  & 26/33 & 0.058 & 0.660 & \textbf{41/46} & 0.042 & 0.444 & 9/13  & 0.0429 & 0.460 \\
    GMR    & 25/33 & 0.048 & 0.244 & 40/46 & 0.046 & 0.450 & 9/13  & 0.043  & 0.319 \\
    NMR (Ours) & \textbf{31/33} & \textbf{0.040} & \textbf{0.237} & \textbf{41/46} & \textbf{0.035} & \textbf{0.308} & \textbf{10/13} & \textbf{0.040}  & \textbf{0.291} \\
    \bottomrule
  \end{tabular}
\end{table*}

\begin{figure*}[htb]
	\centering
	\includegraphics[width=\linewidth]{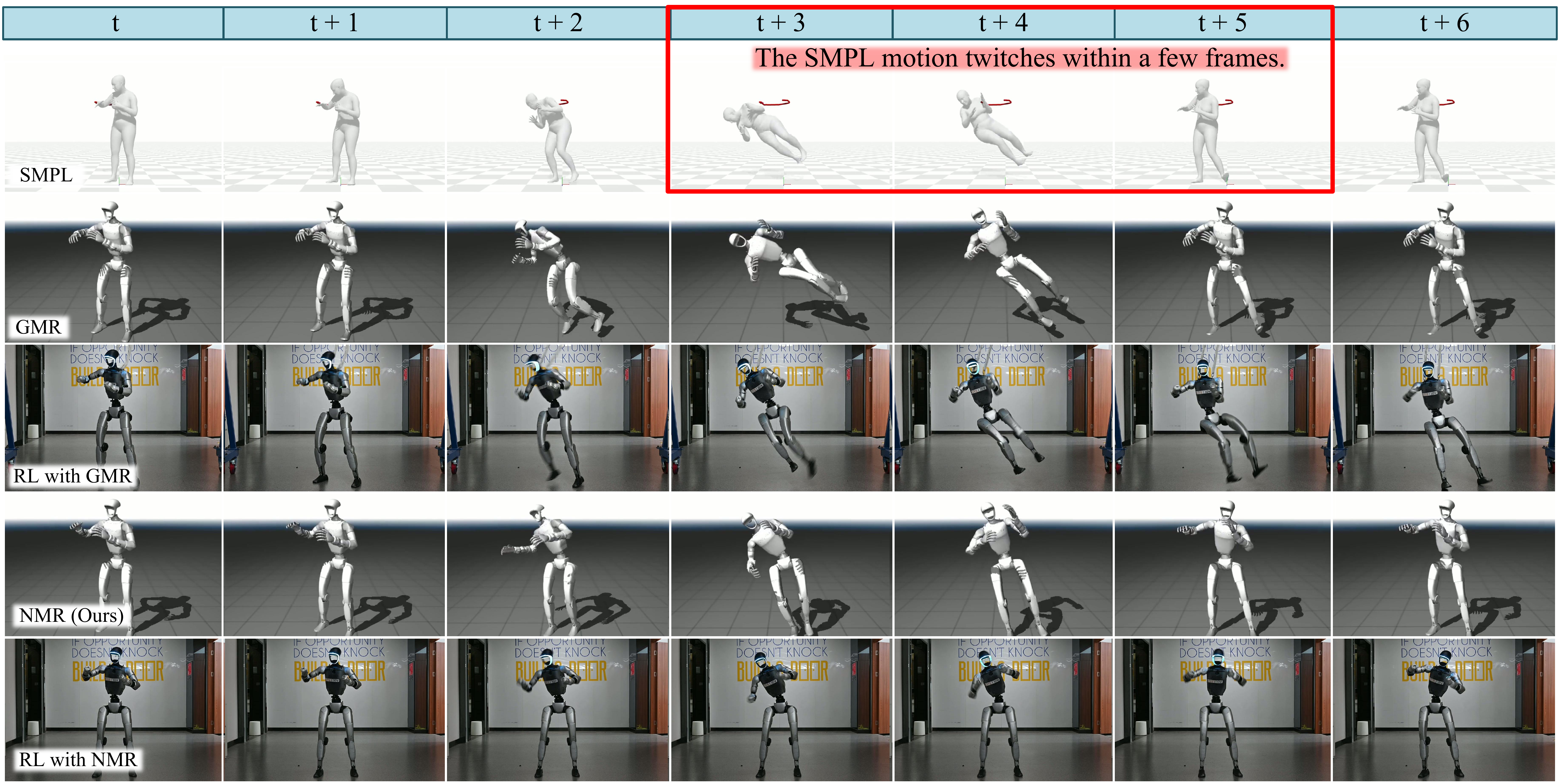}
	\caption{Comparison under abnormal SMPL motions, frame interval is around 0.06 s. When abnormal poses appear in the original SMPL sequence between t+2 and t+5, the proposed NMR method implicitly filter out the abnormal, producing smoother and more feasible robot motion. In contrast, GMR retain the error from source motion, making it difficult for the learned policy to compensate and causing instability on the real robot.}
 \label{fig:correct_upstream}
\end{figure*}

We evaluate retargeted motions on three physical plausibility metrics.
\textbf{Joint Jump} counts frames where max single-step joint angle change exceeds 0.5\,rad.
\textbf{Self-Collision} flags frames with non-hand body-segment contacts via MuJoCo forward kinematics.
\textbf{Joint Limit} counts frames where any joint comes within 0.05\,rad of its hardware boundary.

As shown in Tab.~\ref{tab:quant_cmp}, NMR achieves the best results across all metrics: zero joint jumps, 54\% fewer self-collisions than GMR, and joint limit violations reduced to 16.80\%—roughly half of PHUMA. The ablation (NMR w/o RL) confirms that physics refinement is essential for hardware-feasible output.

As shown in Figure \ref{fig:visual cmp 1}, The GMR method (green curve) encounters the lower joint limit of the right shoulder roll joint during the initial phase of the motion, causing the optimizer to converge to a local optimum due to improper initialization. During the interval from t=0.4s to t=0.8s, the joint angle of GMR remains stagnant near the limit, resulting in substantial deviation of the robot motion from the original SMPL reference, manifested as distorted upper-limb postures. Once the accumulated error exceeds a certain threshold, GMR abruptly escapes from the local minimum at approximately t=0.8s, with the joint angle undergoing a sudden change of approximately 1.5 rad within 0.2s. The corresponding angular velocity reaches 7.5 rad/s. By contrast, both NMR (blue curve) and PHUMA (orange curve) generate smooth and continuous joint trajectories, with NMR achieving the optimal motion smoothness while maintaining motion similarity.

The downstream effect of motion quality is further reflected in the RL tracking policy training curves (Fig.~\ref{fig:train_log}). Policies trained on NMR references reach longer episodes and higher rewards, indicating that cleaner reference motions yield more efficient policy learning.

Table ~\ref{tab:tracking} reports the end-to-end tracking performance of policies trained on each method's retargeted motions, evaluated by success rate, mean per-joint pose error (MPJPE), and mean per-joint pose error under world frame (W-MPJPE) across short, medium, and long sequences. NMR achieves the highest success rate and lowest MPJPE/W-MPJPE in all settings. Notably, PHUMA's high W-MPJPE on short sequences (0.660\,m vs.\ 0.237\,m for NMR) suggests its sequence-level optimization distorts short motions.

\subsection{Correcting Upstream Errors}
As shown in Figure \ref{fig:correct_upstream}, when the original SMPL sequence contains abnormal jitters caused by pose estimation errors, NMR can implicitly filter out these artifacts and generate smooth, physically feasible robot motion trajectories. 

This robustness arises from two key design choices. First, because our training data have undergone physical-consistency filtering (see Section III-B), motion jitters caused by upstream estimation errors are treated as out-of-distribution samples with respect to the training distribution. Unlike optimization-based methods that solve the problem independently on a frame-by-frame basis, neural networks naturally exhibit a smoothing generalization effect on out-of-distribution inputs: when the input deviates from the training distribution, the model tends to predict continuous and plausible motion sequences rather than mechanically reproducing the input noise. 

Second, our network employs a bidirectional self-attention mechanism, enabling the prediction at each frame to access the full temporal context. This global modeling capability allows the network to leverage the normal motion information before and after an anomalous frame for implicit interpolation and correction, thereby suppressing the propagation of local noise. In contrast, optimization-based methods such as GMR can process only one frame at a time and lack cross-frame constraints on physical plausibility. 

As a result, they tend to pass upstream errors directly to the output, making them difficult for downstream tracking policies to compensate for and thereby causing unstable motion on real robots.

% 因为我们的数据经过了过滤，训练数据中不存在这种由于姿态估计不准导致的动作抽搐，bad motion 属于离群数据，与基于优化的方法相比，神经网络对这种离群数据有很好的泛化能力。模型倾向于预测连续，合理的 motion。并且我们的网络采用了 self attn，并不是单帧优化，每一帧都可以看到全局的 motion，

\section{Conclusion} 
\label{sec:conclusion}
This paper addresses the physical-feasibility gap in human-to-humanoid motion retargeting by proposing NMR, a Neural Motion Retargeting framework. The central insight is to treat retargeting  as a learned distribution mapping rather than a frame-wise geometric optimization. 

To enable this data-driven approach, we propose CEPR, a pipeline that leverages VAE-based clustering and massively parallel RL expert policies to generate approximately 30K physically validated human-robot motion pairs. A Transformer-based network, pretrained on large-scale kinematic data and fine-tuned with CEPR-refined pairs, performs efficient inference without requiring a physics simulator. Experiments on the Unitree G1 demonstrate that NMR achieves zero joint jumps, reduces self-collision frames by 54\%, and cuts joint-limit violations by 61\%, while also accelerating the convergence of downstream whole-body control policies. Furthermore, global temporal attention enables NMR to implicitly suppress upstream SMPL estimation errors rather than propagating them. 

A current limitation is that CEPR is morphology-specific; extending to other platforms requires regenerating the data pipeline, and developing morphology-conditioned architectures remains a direction for future work.

\section*{Acknowledgments}
This research is supported by HUAWEI’s Al Hundred Schools Program and was carried out using the Ascend AI technology stack. Thank Tianhao Jiang for helping with hardware experiment. 
%% Use plainnat to work nicely with natbib. 

\bibliographystyle{IEEEtran}
\bibliography{references}

\section{Appendix}
\subsection{Non-convexity Analysis of Retargeting Optimization}~\label{app:nonconvexity}

This appendix provides the detailed derivation for the non-convexity analysis summarized in Section~\ref{sec: problem_formulation}.

\noindent\textbf{Geometric Preliminaries}

We briefly introduce the notation on the special Euclidean group $SE(3)=SO(3)\ltimes\mathbb{R}^3$ and its Lie algebra $\mathfrak{se}(3)\cong\mathbb{R}^6$.
For any vector $\mathbf{v}\in\mathbb{R}^3$, let $[\mathbf{v}]_\times$ denote the $3\times 3$ skew-symmetric matrix such that
$[\mathbf{v}]_\times \mathbf{u} = \mathbf{v}\times \mathbf{u}$.
For a rotation matrix $R\in SO(3)$, its logarithm is written as
\[
\Log(R)=\boldsymbol{\omega}=\phi \hat{\mathbf{n}},
\]
where $\phi\in[0,\pi)$ is the rotation angle and $\hat{\mathbf{n}}$ is the unit rotation axis.

The left Jacobian of $SO(3)$ and its inverse are
\begin{equation}
\begin{aligned}
    \mathcal{J}_{SO(3)}(\boldsymbol{\omega})
    &= \frac{\sin\phi}{\phi}I
    + \left(1-\frac{\sin\phi}{\phi}\right)\hat{\mathbf{n}}\hat{\mathbf{n}}^\top
    + \frac{1-\cos\phi}{\phi}[\hat{\mathbf{n}}]_\times, \\
    \mathcal{J}_{SO(3)}^{-1}(\boldsymbol{\omega})
    &= \frac{\phi/2}{\tan(\phi/2)}I
    + \left(1-\frac{\phi/2}{\tan(\phi/2)}\right)\hat{\mathbf{n}}\hat{\mathbf{n}}^\top
    - \frac{\phi}{2}[\hat{\mathbf{n}}]_\times .
\end{aligned}
\label{eq:so3jac}
\end{equation}

\noindent\textbf{A Unified Surrogate Objective}

To expose the geometric source of non-convexity, we consider a single end-effector pose $T(\boldsymbol{\theta})\in SE(3)$ and a target pose $T^\star\in SE(3)$, and define the relative pose error
\begin{equation}
    E(\boldsymbol{\theta}) = (T^\star)^{-1}T(\boldsymbol{\theta}) \in SE(3).
\end{equation}
We lift this error to the Lie algebra via
\begin{equation}
    \boldsymbol{\xi}(\boldsymbol{\theta})
    = \Log(E(\boldsymbol{\theta}))
    = \begin{pmatrix}
        \boldsymbol{\omega}(\boldsymbol{\theta}) \\
        \mathbf{v}(\boldsymbol{\theta})
      \end{pmatrix}
    \in \mathbb{R}^6.
    \label{eq:xi_app}
\end{equation}
Here $\boldsymbol{\omega}\in\mathbb{R}^3$ is the rotational log-coordinate and $\mathbf{v}\in\mathbb{R}^3$ is the translational log-coordinate. Note that $\mathbf{v}$ is the translational component of the $SE(3)$ logarithm and, in general, is not identical to the Euclidean position error; rather, it provides a geometrically consistent coupled pose representation.

We then study the weighted surrogate cost
\begin{equation}
    \tilde f(\boldsymbol{\theta})
    = \frac{1}{2}\,\boldsymbol{\xi}(\boldsymbol{\theta})^\top
    W\,
    \boldsymbol{\xi}(\boldsymbol{\theta}),
    \qquad
    W=\mathrm{diag}(w_R I_3,\; w_p I_3),
    \label{eq:surrogate_cost_app}
\end{equation}
with weights $w_R,w_p>0$. This objective is not globally identical to the original retargeting loss, but serves as a geometrically meaningful surrogate for analyzing second-order structure.

\noindent\textbf{Gradient and Hessian Decomposition}

Let
\[
J_{\xi}(\boldsymbol{\theta}) = \frac{\partial \boldsymbol{\xi}}{\partial \boldsymbol{\theta}}
\in \mathbb{R}^{6\times n}
\]
denote the Jacobian of the log-coordinate error. By direct differentiation,
\begin{equation}
    \nabla \tilde f(\boldsymbol{\theta})
    = J_{\xi}(\boldsymbol{\theta})^\top W \boldsymbol{\xi}(\boldsymbol{\theta}),
\end{equation}
and the Hessian is
\begin{equation}
    \nabla^2 \tilde f(\boldsymbol{\theta})
    =
    \underbrace{J_{\xi}^\top W J_{\xi}}_{\text{Gauss--Newton term, PSD}}
    +
    \underbrace{\sum_{a=1}^{6}(W\boldsymbol{\xi})_a \,\nabla^2 \xi_a}_{\text{curvature correction}}.
    \label{eq:hessian_decomp_app}
\end{equation}
The first term is always positive semi-definite. Therefore, any negative curvature must arise from the second term, which captures the second-order geometry of the error map $\boldsymbol{\theta}\mapsto \Log((T^\star)^{-1}T(\boldsymbol{\theta}))$.

\noindent\textbf{Source I: Curvature Induced by Forward Kinematics}

Even if one ignores the logarithmic-map nonlinearity and focuses only on translational kinematics, the mapping from joint angles to end-effector position is nonlinear due to recursive trigonometric composition.
Consider a planar or spatial serial chain with at least two revolute joints ($n\ge 2$), and let the target position be chosen so that the translational error is aligned with the fully extended direction of the chain. Around the canonical extended configuration $\boldsymbol{\theta}=\mathbf{0}$, the end-effector position admits the standard second-order expansion
\begin{equation}
    p(\boldsymbol{\theta})
    =
    p(\mathbf{0})
    + J_p(\mathbf{0})\,\boldsymbol{\theta}
    + \frac{1}{2}\,\mathcal{H}_p(\mathbf{0})[\boldsymbol{\theta},\boldsymbol{\theta}]
    + o(\|\boldsymbol{\theta}\|^2),
\end{equation}
where $J_p$ is the translational Jacobian and $\mathcal{H}_p$ denotes the second derivative tensor of forward kinematics.

For a perturbation direction that bends an interior joint away from the extended pose, the second-order positional variation points opposite to the extension direction. Hence, if the target is placed further along the extension direction, the translational component of the curvature correction contributes negatively along that perturbation. Equivalently, there exists a direction $\mathbf{u}\in\mathbb{R}^n$ and a target pose $T^\star$ such that
\begin{equation}
    \mathbf{u}^\top
    \left(
    \sum_{a=4}^{6}(W\boldsymbol{\xi})_a \nabla^2 \xi_a
    \right)
    \mathbf{u}
    <0
\end{equation}
at $\boldsymbol{\theta}=\mathbf{0}$.
Thus, the nonlinear forward-kinematics map alone can induce negative second-order curvature.

\noindent\textbf{Source II: Curvature Induced by the Logarithmic Map on $SO(3)$}

A second and geometrically distinct source of non-convexity comes from the differential of the logarithm map itself.
Consider the pure rotational component
\[
\boldsymbol{\omega}(\boldsymbol{\theta})=\Log(R^\star{}^\top R(\boldsymbol{\theta})).
\]
Its Jacobian involves the inverse left Jacobian $\mathcal{J}_{SO(3)}^{-1}(\boldsymbol{\omega})$, whose coefficients depend nonlinearly on the rotation angle $\phi=\|\boldsymbol{\omega}\|$.
In particular, the scalar coefficient
\begin{equation}
    \alpha(\phi)=\frac{\phi/2}{\tan(\phi/2)}
\end{equation}
satisfies
\begin{equation}
    \alpha'(\phi)<0,
    \qquad \phi\in(0,\pi),
\end{equation}
which shows that the differential of $\Log$ varies nonlinearly and increasingly sharply as the rotation error approaches $\pi$.

Now choose a target orientation $R^\star$ such that at some configuration $\boldsymbol{\theta}_0$ the relative rotation
\[
R^\star{}^\top R(\boldsymbol{\theta}_0)
\]
has angle $\phi(\boldsymbol{\theta}_0)\in(\pi/2,\pi)$, and choose a perturbation direction $\mathbf{u}$ such that $\mathrm{d}\phi(\boldsymbol{\theta}_0)[\mathbf{u}]\neq 0$.
Then the second derivative of the rotational log error contains a term proportional to the variation of $\mathcal{J}_{SO(3)}^{-1}$ with respect to $\phi$, which contributes negatively along $\mathbf{u}$ for a suitable choice of target and local motion direction. Therefore, there exist $\boldsymbol{\theta}_0$, $R^\star$, and $\mathbf{u}$ such that
\begin{equation}
    \mathbf{u}^\top
    \left(
    \sum_{a=1}^{3}(W\boldsymbol{\xi})_a \nabla^2 \xi_a
    \right)
    \mathbf{u}
    <0 .
\end{equation}
This shows that negative curvature can arise even when the kinematic map is locally regular, purely due to the geometry of the rotational logarithm.

\noindent\textbf{Proof of Proposition 1}

The two mechanisms above are geometrically distinct: the first originates from the second-order curvature of forward kinematics, while the second comes from the nonlinearity of the logarithmic chart on $SO(3)$/$SE(3)$.
Together they imply the following result.

\medskip
\noindent\textbf{Proposition 1 (existence of negative curvature).}
\emph{
Let $n\ge 2$ and $w_R,w_p>0$. For the surrogate objective $\tilde f$ in~\eqref{eq:surrogate_cost_app}, there exist feasible target poses $T^\star$ and robot configurations $\boldsymbol{\theta}$ such that the Hessian $\nabla^2 \tilde f(\boldsymbol{\theta})$ has a strictly negative directional curvature; namely, there exists a direction $\mathbf{u}\neq \mathbf{0}$ satisfying
\begin{equation}
    \mathbf{u}^\top \nabla^2 \tilde f(\boldsymbol{\theta}) \mathbf{u}<0.
\end{equation}
Hence, $\tilde f$ is generally non-convex.
}

\begin{proof}
By the analysis in Sections~\ref{app:nonconvexity}.4 and~\ref{app:nonconvexity}.5, there exist configurations where either the forward-kinematics curvature or the logarithmic-map curvature dominates, making the curvature correction term in~\eqref{eq:hessian_decomp_app} sufficiently negative to overcome the positive semi-definite Gauss--Newton term along certain directions. This establishes the existence of negative directional curvature.
\end{proof}

\subsection{Test Motion Names Used in Evaluation}
% 左侧主表（ULOM + 部分WBCM）
\begin{minipage}[t]{0.48\textwidth}
\centering
\scriptsize  % 紧凑字体，适配分栏
\captionof{table}{List of Test Motion Files from AMASS (1/2)}
\label{tab:amass_motion_1}
\begin{tabular}{l l l}
\toprule
\textbf{Type} & \textbf{Source} & \textbf{File Name} \\
\midrule
% ULOM 类型
ULOM/L   & ACCAD (Female1Gestures\_c3d) & D2 \\
ULOM/L   & ACCAD (Female1Gestures\_c3d) & D3 \\
ULOM/L   & CMU                         & 02\_07 \\
ULOM/L   & CMU                         & 02\_10 \\
ULOM/L   & CMU                         & 13\_20 \\
ULOM/M   & ACCAD (Female1Gestures\_c3d) & D1 \\
ULOM/M   & ACCAD (Male1General\_c3d)    & General\_A2 \\
ULOM/M   & ACCAD (Male1General\_c3d)    & General\_A4 \\
ULOM/M   & ACCAD (Male1General\_c3d)    & General\_A5 \\
ULOM/M   & ACCAD (Male1General\_c3d)    & General\_A6 \\
ULOM/M   & CMU                         & 02\_05 \\
ULOM/M   & CMU                         & 02\_08 \\
ULOM/M   & CMU                         & 02\_09 \\
ULOM/M   & CMU                         & 79\_46 \\
ULOM/M   & CMU                         & 79\_90 \\
ULOM/S   & ACCAD (Male2MartialArtsPunches\_c3d) & E1 \\
ULOM/S   & ACCAD (Male2MartialArtsPunches\_c3d) & E2 \\
ULOM/S   & ACCAD (Male2MartialArtsPunches\_c3d) & E3 \\
ULOM/S   & ACCAD (Male2MartialArtsPunches\_c3d) & E5 \\
ULOM/S   & ACCAD (Male2MartialArtsPunches\_c3d) & E6 \\

% WBCM 类型（前半部分）
WBCM/L   & ACCAD (Male2MartialArtsExtended\_c3d) & Extended\_1 \\
WBCM/L   & ACCAD (Male2MartialArtsExtended\_c3d) & Extended\_2 \\
WBCM/L   & ACCAD (Male2MartialArtsExtended\_c3d) & Extended\_3 \\
WBCM/L   & CMU                         & 01\_01 \\
WBCM/L   & CMU                         & 06\_13 \\
WBCM/L   & CMU                         & 122\_22 \\
WBCM/M   & ACCAD (Female1Gestures\_c3d) & D5 \\
WBCM/M   & ACCAD (Male2MartialArtsStances\_c3d)  & D10 \\
WBCM/M   & ACCAD (Male2MartialArtsStances\_c3d)  & D6 \\
WBCM/M   & ACCAD (Male2MartialArtsStances\_c3d)  & D7 \\
WBCM/M   & ACCAD (Male2MartialArtsStances\_c3d)  & D9\_-\_t2 \\
WBCM/M   & ACCAD (Male2MartialArtsStances\_c3d)  & D9\_-\_victory \\
WBCM/M   & ACCAD (Male2MartialArtsStances\_c3d)  & D9\_-\_warm \\
\bottomrule
\end{tabular}
\end{minipage}

\hfill % 自动填充左右栏间距，保证对称
% 右侧续表（剩余WBCM + 全部WBPM）
\begin{minipage}[t]{0.48\textwidth}
\centering
\scriptsize  % 与左侧字体一致
\captionof{table}{List of Test Motion Files from AMASS (2/2, continue)}
\label{tab:amass_motion_2}
\begin{tabular}{l l l}
\toprule
\textbf{Type} & \textbf{Source} & \textbf{File Name} \\
\midrule
WBCM/M   & CMU                         & 05\_02 \\
WBCM/M   & CMU                         & 05\_04 \\
WBCM/M   & CMU                         & 05\_18 \\
WBCM/M   & CMU                         & 106\_15 \\
WBCM/S   & ACCAD (Male2MartialArtsKicks\_c3d)    & G11 \\
WBCM/S   & ACCAD (Male2MartialArtsKicks\_c3d)    & G16 \\
WBCM/S   & ACCAD (Male2MartialArtsKicks\_c3d)    & G17 \\
WBCM/S   & ACCAD (Male2MartialArtsKicks\_c3d)    & G19 \\
WBCM/S   & ACCAD (Male2MartialArtsStances\_c3d)  & D13 \\
WBCM/S   & ACCAD (Male2MartialArtsStances\_c3d)  & D14 \\
WBCM/S   & ACCAD (Male2MartialArtsStances\_c3d)  & D1 \\
WBCM/S   & ACCAD (Male2MartialArtsStances\_c3d)  & D3 \\
WBCM/S   & CMU                         & 02\_04 \\

% WBPM 类型（全部）
WBPM/L   & CMU                         & 02\_06 \\
WBPM/L   & CMU                         & 137\_03 \\
WBPM/M   & ACCAD (Female1Running\_c3d) & C10 \\
WBPM/M   & ACCAD (Female1Running\_c3d) & C14 \\
WBPM/M   & ACCAD (Female1Running\_c3d) & C21 \\
WBPM/M   & ACCAD (Female1Running\_c3d) & C25 \\
WBPM/M   & ACCAD (Female1Running\_c3d) & C2 \\
WBPM/M   & ACCAD (Female1Running\_c3d) & C5 \\
WBPM/M   & ACCAD (Female1Running\_c3d) & C6 \\
WBPM/M   & ACCAD (Female1Running\_c3d) & C8 \\
WBPM/M   & ACCAD (Female1Walking\_c3d) & B11 \\
WBPM/M   & ACCAD (Female1Walking\_c3d) & B15 \\
WBPM/M   & ACCAD (Female1Walking\_c3d) & B16 \\
WBPM/M   & ACCAD (Female1Walking\_c3d) & B20 \\
WBPM/M   & ACCAD (Female1Walking\_c3d) & B21 \\
WBPM/M   & ACCAD (Female1Walking\_c3d) & B21\_s2 \\
WBPM/M   & ACCAD (Female1Walking\_c3d) & B21\_s3 \\
WBPM/M   & ACCAD (Female1Walking\_c3d) & B22 \\
WBPM/M   & ACCAD (Female1Walking\_c3d) & B5 \\
WBPM/M   & ACCAD (Male1General\_c3d)   & General\_A12 \\
WBPM/M   & CMU                         & 06\_10 \\
WBPM/M   & CMU                         & 06\_12 \\
WBPM/M   & CMU                         & 09\_12 \\
WBPM/M   & CMU                         & 122\_33 \\
WBPM/M   & CMU                         & 122\_36 \\
WBPM/M   & CMU                         & 137\_04 \\
WBPM/M   & CMU                         & 143\_38 \\
WBPM/S   & ACCAD (Female1Running\_c3d) & C13 \\
WBPM/S   & ACCAD (Female1Running\_c3d) & C21\_s2 \\
WBPM/S   & ACCAD (Female1Running\_c3d) & C26 \\
WBPM/S   & ACCAD (Female1Running\_c3d) & C27 \\
WBPM/S   & ACCAD (Female1Running\_c3d) & C4 \\
WBPM/S   & ACCAD (Female1Running\_c3d) & C7 \\
WBPM/S   & CMU                         & 02\_01 \\
WBPM/S   & CMU                         & 02\_02 \\
WBPM/S   & CMU                         & 02\_03 \\
WBPM/S   & CMU                         & 07\_01 \\
WBPM/S   & CMU                         & 07\_02 \\
WBPM/S   & CMU                         & 07\_03 \\
WBPM/S   & CMU                         & 07\_06 \\
WBPM/S   & CMU                         & 07\_07 \\
WBPM/S   & CMU                         & 07\_08 \\
WBPM/S   & CMU                         & 07\_09 \\
WBPM/S   & CMU                         & 07\_10 \\
WBPM/S   & CMU                         & 07\_11 \\
WBPM/S   & CMU                         & 07\_12 \\
\bottomrule
\end{tabular}
\end{minipage}

\vspace{1.5em} % 全局底部间距，保留下边距，避免贴页底

\end{document}